\documentclass[runningheads]{llncs}

\usepackage[T1]{fontenc}
\usepackage{graphicx}
\usepackage{multicol}
\usepackage{amssymb}
\usepackage{multirow}
\usepackage{tabularx}
\usepackage{arydshln}
\usepackage{amsmath}
\usepackage{amssymb}
\usepackage{tablefootnote}
\usepackage{bbold}
\usepackage{wrapfig}
\usepackage{url}
\usepackage[dvipsnames]{xcolor}
\usepackage{subcaption}

\def\ours{\texttt{\textbf{SITAR}}}

\definecolor{myGreen}{HTML}{008000}

\begin{document}
\title{\ours: Semi-supervised Image Transformer for Action Recognition}

\titlerunning{\ours: Semi-supervised Image Transformer for Action Recognition}
\author{Owais Iqbal\inst{1} \and
Omprakash Chakraborty\inst{1} \and
Aftab Hussain\inst{1} \and
Rameswar Panda\inst{2} \and
Abir Das\inst{1}}
\authorrunning{O. Iqbal et al.}

\institute{$^1$ IIT-Kharagpur, India, 
	$^2$ MIT-IBM Watson AI Lab\\
	\email{\{owais.iqbal@kgpian, opckgp@, aftabh4004@kgpian, abir@cse\}.iitkgp.ac.in, rpanda@ibm.com}}%

\maketitle
\begin{abstract}
Recognizing actions from a limited set of labeled videos remains a challenge as annotating visual data is not only tedious but also can be expensive due to classified nature. Moreover, handling spatio-temporal data using deep $3$D transformers for this can introduce significant computational complexity. In this paper, our objective is to address video action recognition in a semi-supervised setting by leveraging only a handful of labeled videos along with a collection of unlabeled videos in a compute efficient manner. Specifically, we rearrange multiple frames from the input videos in row-column form to construct super images. Subsequently, we capitalize on the vast pool of unlabeled samples and employ contrastive learning on the encoded super images. Our proposed approach employs two pathways to generate representations for temporally augmented super images originating from the same video. Specifically, we utilize a $2$D image-transformer to generate representations and apply a contrastive loss function to minimize the similarity between representations from different videos while maximizing the representations of identical videos. Our method demonstrates superior performance compared to existing state-of-the-art approaches for semi-supervised action recognition across various benchmark datasets, all while significantly reducing computational costs.
\footnote{Project page: \url{https://cvir.github.io/projects/sitar}}

\end{abstract}

\section{Introduction}

Video action recognition (VAR) in computer vision remains a long-standing challenge~\cite{brand1997coupled,gaur2011string,laxton2007leveraging,tran2015learning,veeraraghavan2009rate}. 
Earlier approaches relied on optical flow~\cite{ke2007spatio,simonyan2014two} or $3$D convolutions using spatio-temporal kernels~\cite{tran2015learning} for feature extraction.
However, recent advancements have introduced transformer architectures~\cite{arnab2021vivit,Fan_2021_ICCV} to address this task.
A key challenge in achieving high accuracy for VAR lies in the scarcity of large-scale, meticulously labeled videos.
The hunger for annotated data has further aggravated with the advent of $3$D video transformers~\cite{arnab2021vivit} which continuously surpass the performance by $2$D vision transformers but requires significantly higher amount of labeled videos to perform. At the same time, $3$D Transformers have more parameters, making it computationally more inefficient compared to ConvNets or $2$D transformers.
Though carefully annotating videos is expensive and labor-intensive, with the proliferation of video cameras and internet millions of videos without labels are available in the wild.

In scenarios with limited labeled training data alongside a substantial amount of unlabeled samples,
semi-supervised learning (SSL)~\cite{yang2022survey} has shown significant promise.
Despite the inherent promise of semi-supervised video action recognition, it remains a relatively under-explored area compared to its fully supervised counterpart.
A straightforward approach of applying image based semi-supervised learning techniques to videos might not be sufficient to bridge the performance gap. 
The reason being the additional temporal dimension in videos.
However, if properly used, \emph{temporal information} can be a friend instead of a foe to semi-supervised video action recognition.
Barring a few canonical activities (\textit{e.g.}, walking vs jogging) actions do not change if a video is played fast or slow.
A robust video action recognition system should have the capability to identify actions regardless of the speed of the video.
Recent works~\cite{Feichtenhofer_2019_ICCV,xiao2020audiovisual} demonstrate the effectiveness of training models to be invariant across different play-rates of the same actions.
However, these approaches are still supervised and utilise Convolutional Neuron Networks (CNNs) to process the videos.
Learning invariances among different versions of same samples as a means of self-supervision has been showing promising results with the emergence of contrastive learning. 
Studies such as~\cite{chen2020simple,he2020momentum} have exhibited superior performance compared to supervised learning approaches in image classification.
This concept has been extended to videos~\cite{qian2021spatiotemporal,Dorkenwald_2022_CVPR}, where~\cite{qian2021spatiotemporal} utilizes temporally distant clips from a video alongside spatial augmentations as positives, while negatives are selected from different videos.

Inspired by the success of utilizing both slow and fast versions of video for supervised action recognition and the success of contrastive learning frameworks~\cite{qian2021spatiotemporal,pan2021videomoco}, this work presents a new approach for semi-supervised video action recognition (VAR) that leverages unlabeled video data in both computationally and label efficient manner. 
Our method differs from traditional approaches by first rearranging video frames into informative super images~\cite{fan-iclr2022}.
Our approach initially transforms a $3$D video into a $2$D image by rearranging a sequence of input video frames into a super image based on a predetermined spatial layout.
These super images capture both spatial and temporal information within video segments, enabling the model to learn representations efficiently. Like images, with super images too, an image transformer can be used. 

To leverage temporal information, we introduce a two-pathway model.  Each pathway processes temporally augmented versions of the same video (\textit{e.g.}, slow and fast versions).
In the fast pathway, the super image is formed using double the number of frames compared to the super image in the slow pathway.
Despite this variation, these super images share the same semantic content when they come from the same unlabeled video. 
The model learns to maximize this semantic similarity. At the same time, the model tries to minimize the similarity between the representations coming from the super images of two different videos in these two pathways.
It is achieved by using a contrastive loss~\cite{chen2020simple} formulation.
We extensively experiment with three benchmark datasets employing the Swin Transformer~\cite{liu2021swin} as the backbone.
Our approach \textbf{S}emi-Supervised \textbf{I}mage \textbf{T}ransformer for
\textbf{A}ction \textbf{R}ecognition (\ours), demonstrates superior performance outperforming previous approaches. Additionally, \ours \ also exhibits better computational and parameter efficiency compared to existing semi-supervised video action recognition approaches.

\section{Related Works}

\textbf{Video Action Recognition.} Convolutional Neural Networks (CNNs)~\cite{feichtenhofer2020x3d,Feichtenhofer_2019_ICCV,tran2018closer} and transformers~\cite{arnab2021vivit,Fan_2021_ICCV,li2022mvitv2} have emerged as dominant approaches for action recognition in videos.
Although, $3$D CNNs~\cite{tran2018closer,Feichtenhofer_2019_ICCV,feichtenhofer2020x3d} have been dominant in earlier works, their large number of parameters necessitates vast training datasets. Additionally, CNNs limited receptive field hinders accurate motion modeling, hindering performance. Drawing inspiration from the success of transformers in natural language processing, vision-transformers based architectures like TimeSformer~\cite{bertasius2021spacetime}, MViT~\cite{Fan_2021_ICCV}  and MViTv2~\cite{li2022mvitv2} have emerged as powerful alternatives achieving state-of-the-art performance and also efficient arcitectures~\cite{ryali2023hiera}.
Recent works also underscore the ability of transformers~\cite{tong2022videomae,svformer} on fewer data.
However, these existing video action recognition approaches are supervised and rely on $3$D operations greatly increasing the compute complexity. 
Our work addresses this challenge by utilizing a $2$D Image Transformer in a semi-supervised setting for video action recognition. 
This strategy aims to not only achieve better performance but also significantly lower the computational requirements.

\noindent\textbf{Semi-supervised learning.} While extensive research has yielded successful semi-supervised learning methods in images, directly applying them to VAR proves suboptimal due to the non-exploration of temporal dynamics inherent in videos. 
Pioneering image-based approaches like Pseudo-Labeling  established a foundation,
which leverages the predicted confidence scores (softmax probabilities) from the model itself to generate pseudo-labels for unlabeled data. 
Subsequently, these pseudo-labels are employed to train the network along with a limited set of labeled data.
Later works focused on improving pseudo-label quality, \textit{e.g.}, UPS~\cite{rizve2021defense}, FixMatch~\cite{sohn2020fixmatch}. 
FixMatch utilizes weakly augmented unlabeled instances to generate pseudo-labels and ensures consistent predictions against their strongly augmented counterparts.
FixMatch showed superior performance 
with its influence extending to detection~\cite{wang20213dioumatch} and segmentation~\cite{zou2021pseudoseg}.

\noindent\textbf{Semi-supervised Video Action Recognition.} Despite the emergence of several semi-supervised video action recognition approaches~\cite{Jing_2021_WACV,singh2021semisupervised,xiong2021multiview,zou2021pseudoseg,dass2024actnetformer}, most of them primarily extend image-based approaches to videos, ignoring the crucial temporal dynamics inherent in video data.
Existing semi-supervised VAR approaches like VideoSSL~\cite{Jing_2021_WACV} replaces a $2$D ResNet with a $3$D ResNet~\cite{hara2018spatiotemporal} but is limited to using pseudo-labels and distillation from an image CNN when it comes to exploiting unlabelled videos.
The marginal gain over pure image-based baselines highlights the limitations of directly transferring image-based methods.
With time, more sophisticated video-specific methods, such as TCL~\cite{singh2021semisupervised} have emerged, focusing on exploiting temporal information of the video.
MvPL~\cite{xiong2021multiview} and LTG~\cite{xiao2022learning} utilize multi-modal data such as optical flow or temporal gradient information, respectively, to generate high-quality pseudo-labels for training the network.  Whereas, CMPL~\cite{xu2022crossmodel} introduces an auxiliary network that requires processing more frames during training, potentially increasing computational complexity.  Additionally, all these prior approaches rely on 2D~\cite{singh2021semisupervised} or $3$D convolutional neural networks~\cite{xiao2022learning,xiong2021multiview,xu2022crossmodel}, which can necessitate longer training times.
The recent SOTA, SVFormer~\cite{svformer}, introduces the exploration of $3$D video transformers for semi-supervised action recognition. It incorporates a robust pseudo-labeling framework, utilizing an Exponential Moving Average (EMA) Teacher network, to effectively manage unlabeled video samples.

However, these approaches are computationally expensive. 
Our method addresses this by constructing informative super images from video frames, enabling efficient contrastive learning with a two-pathway model.
This innovative approach significantly reduces computational costs while achieving superior performance on semi-supervised action recognition tasks.
\section{Methodology}
\subsection{Problem Definition}
This work addresses the challenge of semi-supervised action recognition in videos using an Image Transformer.
The input to the model has two subset -
Labeled Videos ($D_l$) and Unlabeled Videos ($D_u$). The labeled video set ($D_l$), comprises of $N_l$ video-label pairs. Each pair ($V^i$, $y^i$) represents the $i^{th}$ video and its corresponding activity label, where $i$ ranges from $1$ to $N_l$ and each $y^i$ belongs to the label set $Y = \{1, 2, ..., C\}$, representing C distinct action categories.
Unlabeled Videos ($D_u$) comprises a significantly larger set $N_u$ ($\gg N_l$) of videos without labels($U^i$), where $i$ ranges from $1$ to $N_u$.
we create two different versions of each unlabeled video by playing them at two different frame rates: fast and slow.

\subsection{\ours \; Framework}
\label{subsec:framework}
\begin{figure}[!t]
  \includegraphics[width=0.95\textwidth]{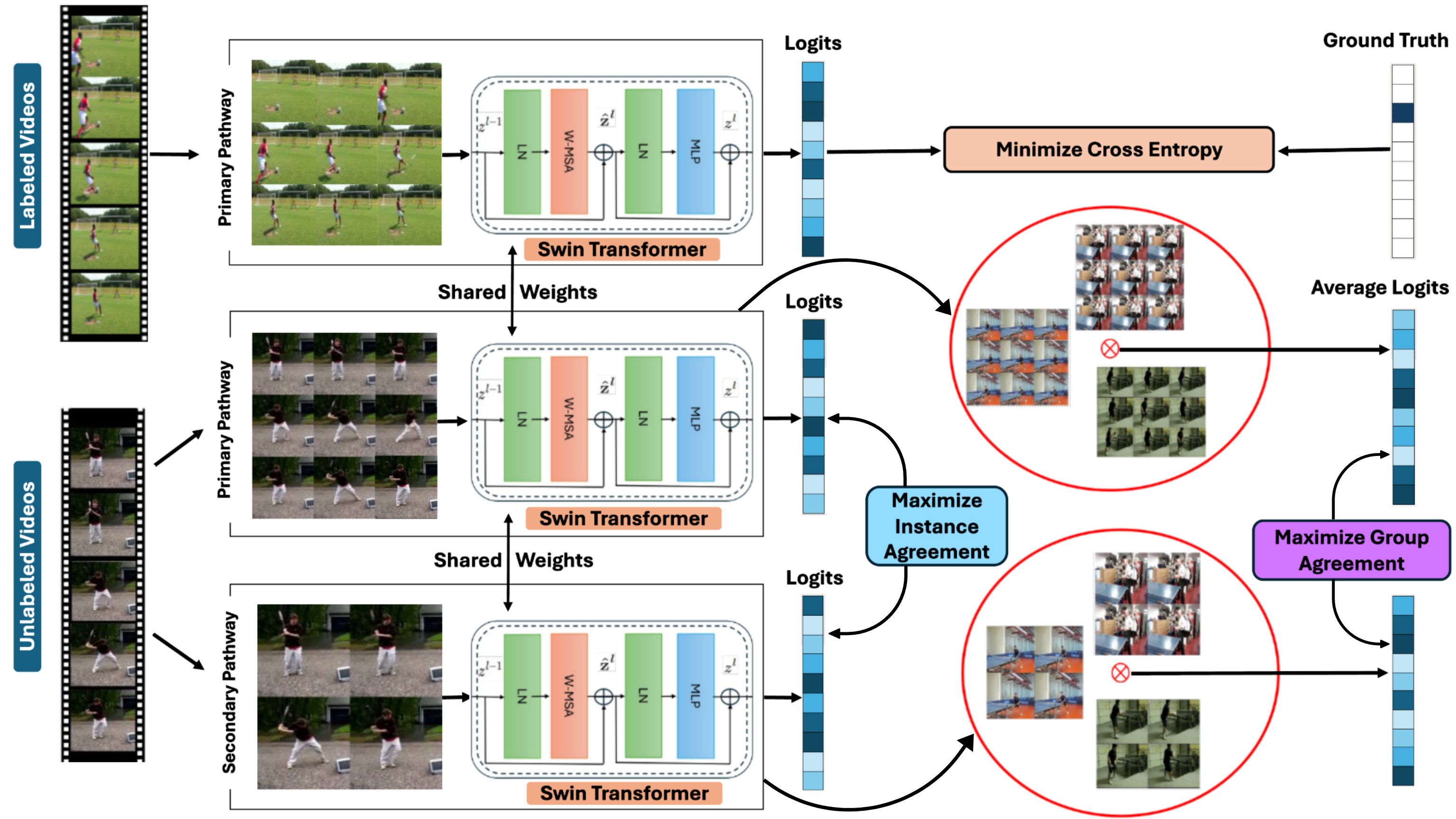}
  \caption{ \textbf{\ours \: framework:} The proposed framework uses two pathways to process the unlabeled videos, namely, \textit{Primary} and \textit{Secondary}, using an image-transformer backbone and sharing the same weights. The Primary pathway is initially trained with limited labeled data. Then, we generate two versions of super images for the unlabeled videos, one fast, having more frames and other slow, having lower frames and pass them through Primary and Secondary pathways respectively. The training objective is to maximize the agreement between the output predictions of the two pathways. To achieve this, we employ two types of contrastive losses. First, an instance contrastive loss to align the representations of a given unlabeled super image across both the pathways. Second, a group contrastive loss to align the average representations of unlabeled super images grouped using pseudo-labels. During inference, only the Primary pathway is used to indentify actions. (Best viewed in color.)}
  \label{fig:model}
  \vspace{-1.5\baselineskip}
\end{figure}

The proposed Semi-Supervised Image Transformer for Action Recognition(\ours) framework, as illustrated in Fig.~\ref{fig:model}, processes video inputs through two distinct pathways: \emph{primary} and \emph{secondary}. The primary pathway handles fast super image $S^i_f$, while the secondary pathway handles slow super image $S^i_s$.
Next we describe how exactly the super images for the two pathways are formed.

\subsubsection{\textbf{Super Image Construction:}}
\begin{wrapfigure}{!hr}{0.4\textwidth}
  \centering
  \includegraphics[width=0.4\textwidth]{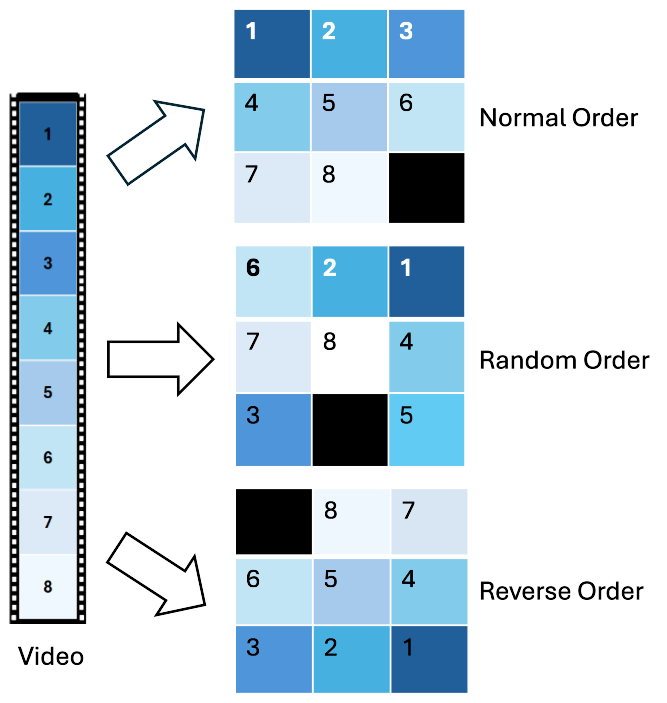}
  \vspace{-2\baselineskip}
  \caption{\small \textbf{Construction of Super Image.} Aligning frames to form a grid using three different temporal orderings: \texttt{normal}, \texttt{random}, and \texttt{reverse}. 
  \label{fig:super_image_order}
}
\vspace{-2\baselineskip}
\end{wrapfigure}
For the primary pathway, each video from $U^i$ is represented as a set of $M$ frames i.e., $U^i_f$ $=$ \{$F^i_{f,1}$, $F^i_{f,2}, \cdots , F^i_{f,M}$\} sampled uniformly from consecutive non-overlapping segments following the approach in~\cite{wang2016temporal}.
Similarly, for the secondary pathway, the same video is represented by $N$ frames ($N < M$ and usually $N=\frac{M}{2}$) as $U^i_s$ $=$ \{$F^i_{s,1}$, $F^i_{s,2}, \cdots, F^i_{s,N}$\}, also sampled uniformly using the same method.
The sampled frames are then transformed into purely $2D$ spatial
patterns, generating super images. Inspired by the work~\cite{fan-iclr2022}, we create informative super images from the input video frames.
This involves arranging the $M$ or $N$ input frames (respectively from the fast ($U^i_f$) or the slow version ($U^i_s$)) in a grid format as shown in Fig.~\ref{fig:super_image_order}.
The grid size $(m \times m)$ is determined by the square root of $M$, ensuring efficient representation. Any remaining empty spaces in the grid are filled with padding images.
The resulting super image for the fast video is denoted as $S^i_f$, while the super image for the slow video is denoted as $S^i_s$.
Both pathways utilize the same image transformer backbone, denoted by g(.).
Next we describe the details of the different training stages within our framework.

\vspace{-2mm}
\subsubsection{Phase $1$: Supervised Learning}
\label{subsubsec:phase1}
Initially, the Swin Transformer~\cite{liu2021swin} (a $2$D Image Transformer backbone) is trained exclusively with the small labeled data set ($D_l$) by passing it through the Primary Pathway (refer Fig.~\ref{fig:model}).
The representation $g(VS^i)$ of the super image $VS^i$ generated from video $V^i$ in our framework is derived from the logits of the Swin Transformer. We minimize the conventional supervised cross-entropy loss $\mathcal{L}$$_{sup}$ on the labeled dataset as follows:
\begin{equation}
\mathcal{L}_{sup} = - \sum_{c=1}^{C} (y^i)_c \log (g(VS^i))_c
\end{equation}

\subsubsection{Phase $2$: Semi-supervised Learning}
\label{subsubsec:phase2}
\vspace{-2mm}
Starting with this initial backbone obtained under limited supervision, our objective is to develop a model capable of leveraging a vast pool of unlabeled videos to enhance activity recognition.
To achieve this, we incorporate instance and group contrastive losses as follows:
\subsubsection{Instance-Contrastive Loss}
\vspace{-2mm}
We utilize the temporal augmentations within unlabeled videos, where each video is transformed into both slow $(S^i_s)$ and fast $(S^i_f)$ super images, and then the pairwise contrastive loss is enforced.
In a minibatch containing $B$ unlabeled videos, the model is trained to align the representation $g(S^i_f)$ of the fast super image of video $U^i$  with $g(S^i_s)$ from the slow super image, forming the positive pair.
The remaining $B - 1$ videos form negative pairs $g(S^i_f)$ and $g(S^k_p)$, where the representation of the $k^{th}$ video can originate from either pathway $(p \in \{f, s\})$. As these negative pairs consists of different videos with distinct content, the representations from different pathways are diverged through the application of contrastive loss $\mathcal{L}$$_{ic}$.
\vspace{-1mm}
\begin{equation}
\label{eq:instance}
\vspace{-4mm}
\mathcal{L}_{ic}(S_f^i, S_s^i)\!\!= \!-\!\!\log\!\frac{h\big(g(S_f^i), \!g(S_s^i)\big)}{\!\!h\big(g(S_f^i),\!g(S_s^i)\!\big)\!+\!\!\!\!\!\!\!\sum\limits_{\substack{k=1\\p\in\{s,f\}}}^{B}\!\!\!\!\!\!\mathbb{1}_{\{k \neq i\}}h\big(g(S_f^i),\!g(S_p^k)\!\big)}
\end{equation}

where, $h(\mathbf{u},\mathbf{v}) = \exp\big(\frac{\mathbf{u}^\top \mathbf{v}}{||\mathbf{u}||_2||\mathbf{v||_2}}/\tau\big)$ represents the exponential of cosine similarity and $\tau$ denotes the temperature hyperparameter.
The instance-contrastive loss is calculated for all positive pairs, namely ($S^i_f$, $S^i_s$) and ($S^i_s$, $S^i_f$), within the minibatch. This loss function aims to minimize similarity not only between different videos within each pathway but also across both pathways.

\subsubsection{Group-Contrastive Loss}
\label{subsubsec:gcl}
\begin{figure*}[!t]
  \centering
  \includegraphics[width=0.8\textwidth]{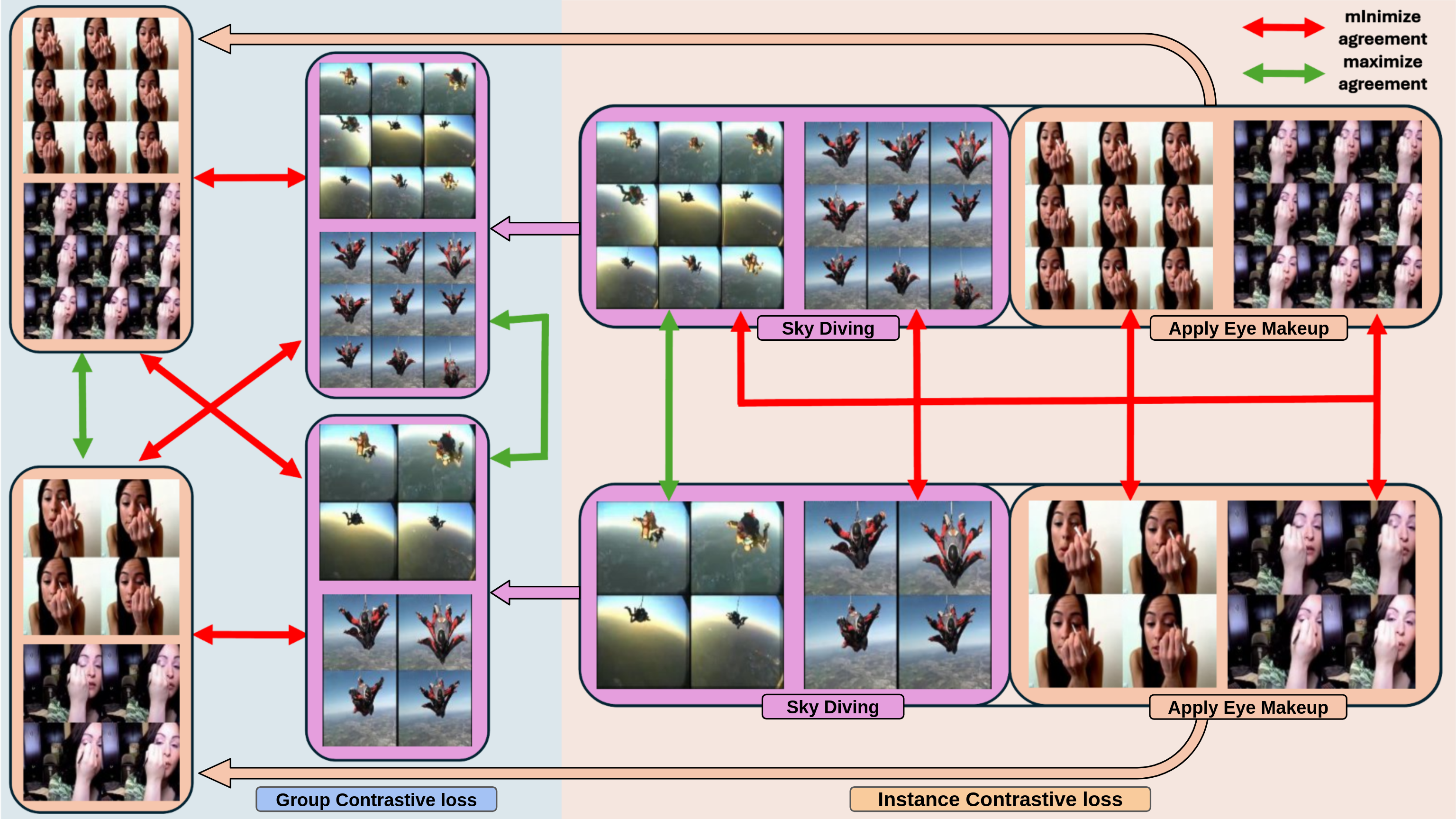}
  \caption{\small \textbf{Instance contrastive loss vs Group contrastive loss.} \ours \: employs two different contrastive losses to leverage on the unlabeled super images. The Instance contrastive loss maximized the agreement between two instances of the same videos which minimizing the agreement with the other videos in a given mini-batch. This risks the same action samples of the mini-batch to be inadvertently pushed apart (right). To mitigate this we employ a Group contrastive loss, which first groups videos with the same activity class (left) as predicted by high-confidence pseudo-labels. Then the average representation is obtained for each group and the contrastive learning policy is applied at this group level. (Best viewed in color.)
  \label{fig:gcl_figure}
}
\vspace{-1.5\baselineskip}
\end{figure*}

\vspace{-2mm}
Standard contrastive loss on unlabeled videos can struggle to capture high-level action semantics without class labels.
As shown in Fig.~\ref{fig:gcl_figure}, it might learn distinct representations for videos with the same action in absence of labels.
To address this, we employ group contrastive loss. This approach uses pseudo-labels assigned to unlabeled videos based on the dominant action class (highest activation).
Let $\hat{y}^i_f$ and $\hat{y}^i_s$ denote the pseudo-labels respectively of the fast ($S^i_f$) and slow ($S^i_s$) super images of the video $U^i$.
Videos with the same pseudo-label within a pathway form a group. The group's representation is the average of its member super image representations (detailed below). This strategy encourages the model to learn more consistent representations for super images with similar actions.
\vspace{-2mm}
\begin{equation}
\label{eq:group_average}
R_p^l = \frac{\sum\limits_{i=1}^{B} \mathbb{1}_{\{\hat{y}_p^i=l\}}g(U_p^i)}{T}
\vspace{-2mm}
\end{equation}

where $\mathbb{1}$ is a binary function that yields a value of $1$ for super images within pathway $p \in \{f, s\}$ whose pseudo-label matches class $l \in Y$. $T$ represents the total number of such super images (with pseudo-label $l$) in pathway $p$ within the minibatch.
Considering the strong agreement between two groups with identical labels in both pathways, it's crucial for these groups to demonstrate comparable features in the feature space. Consequently, within the group-contrastive objective, all pairs $(R^l_f, R^l_s)$ serve as positive pairs, whereas negative pairs consist of pairs $(R^l_f, R^m_p)$ where $p \in \{f, s\}$ and $m \in Y \setminus l$, ensuring that the constituent groups differ in at least one pathway 

\begin{equation}
\hspace{-3mm}\mathcal{L}_{gc}(R_{f}^l, \!R_{s}^l) \!= \!-\!\log\!\frac{h(R_f^l, \!R_s^l)}{h(R_f^l, \!R_s^l)+\sum\limits_{\substack{m=1\\p\in\{s,f\}}}^{C}\!\!\mathbb{1}_{\{m \neq l\}}h(R_f^l, \!R_p^m)}
\vspace{-1mm}
\end{equation}

Like instance-contrastive loss, we compute the group-contrastive loss for all positive pairs $(R^l_f, R^l_s)$ and $(R^l_s, R^l_f)$ over the minibatch. This combined loss, along with the supervised loss $\mathcal{L}_{sup}$ on the labeled data ($D_l$), forms the overall objective function $\mathcal{L}$ used to train our model.

\begin{equation}
\label{eq:total_loss}
\mathcal{L} = \mathcal{L}_{sup} + \gamma \ast \mathcal{L}_{ic} + \beta \ast \mathcal{L}_{gc}
\end{equation}

where, $\gamma$ is the weight assigned to the instance-contrastive loss and $\beta$ is the weight of the group-contrastive loss.

\section{Results and Discussions}
In this section, we first present the experimental settings (Section~\ref{sub:experiment_settings}).
Building on some of the previous works~\cite{xiao2022learning,svformer,xu2022crossmodel}, we show results using different amount of labeled data across three datasets (Section~\ref{sub:main_results}).
Furthermore, we conduct extensive ablation experiments in Section~\ref{sub:ablations}.
For consistency, we utilize the same splits of data as those used by the authors in SVFormer~\cite{svformer}.

\subsection{Experimental Settings}
\label{sub:experiment_settings}
\subsubsection{Datasets.}
Kinetics-$400$~\cite{kay2017ics} is an extensive dataset tailored for human action recognition, containing over $300$K clips distributed across $400$ distinct classes of human actions.
Following the standard practice~\cite{svformer,xiong2021multiview,xu2022crossmodel}, we experimented with two different scenarios in Kinetics-400.
These two scenarios assume only $1\%$ and $10\%$ of all the videos are labeled while the rest are unlabeled.
This amounts to $6$ and $60$ labeled training videos per category in this dataset.
UCF-$101$~\cite{soomro2012ucf101} is another commonly used human action recognition dataset, encompassing a diverse range of categories such as human-human interaction, human-object interaction, playing musical instruments, daily sports \textit{etc}. 
With $13,320$ video samples, UCF-$101$ is spread across $101$ classes. 
Consistent with CMPL~\cite{xu2022crossmodel}, we assume only $1\%$ and $10\%$ of all the videos to be labeled (resulting in $1$ and $10$ labeled samples respectively per category) in this dataset also.
Additionally, HMDB-$51$~\cite{hmdb} comprises human motion data encompassing a variety of actions, including facial expressions combined with object manipulation, general body movements, and movements related to human interaction. 
With $51$ categories and a total of $6,766$ videos, HMDB-$51$ is a comparatively smaller dataset. 
Adhering to the experimental framework established by LTG~\cite{xiao2022learning} and VideoSSL~\cite{Jing_2021_WACV}, we explore three scenarios assuming $40\%$, $50\%$, and $60\%$ of the data to be labeled.

\subsubsection{Baselines.} We benchmark our method against several baselines and existing semi-supervised VAR approaches. 
Initially, we investigate a supervised baseline where we report the performance of training a $3$D-ResNet-$50$ only on the labeled data.
This acts as the lower bound of performance.
Second, we compare with state-of-the-art semi-supervised learning approaches, including FixMatch (NeurIPS'20)~\cite{sohn2020fixmatch}, VideoSSL (WACV'21)~\cite{Jing_2021_WACV}, TCL (CVPR'21)~\cite{singh2021semisupervised}, ActorCutMix (CVIU'21)~\cite{zou2022learning}, MvPL (ICCV'21)~\cite{xiong2021multiview}, CMPL (CVPR'22)~\cite{xu2022crossmodel}, LTG (CVPR'22)~\cite{xiao2022learning}, TACL (TCSVT'22)~\cite{9904603}, L2A (ECCV'22)~\cite{gowda2022learn2augment} and SVFormer (CVPR'23)~\cite{svformer}.
Notably, most of these are computation heavy as $3$D videos are processed, unlike our method, which solely relies on $2$D Image processing.
\vspace{-3mm}
\subsubsection{Implementation Details.} 

We employ uniform sampling to generate video inputs for our models, dividing each video into multiple segments of equal length.
In \ours, we employ uniformly sampled $8$-frame segments for the fast super image and $4$-frame segments for the slow super image.
These images are then fed into the primary and secondary pathways, respectively.
The super image size remains constant at $(572 \times 572)$ for both the primary and auxiliary pathways. As a result, the frame size in the fast and slow super images is adjusted to 192 and 288 each side, respectively.
Note that initially when the model is trained only with the handful of labeled videos, we pass $8$-frame super images from them via the primary pathway.
We start with Swin-B and Swin-S backbones~\cite{liu2021swin} pretrained on ImageNet-$21K$~\cite{5206848}.
Initially, we apply random scaling and cropping for data augmentation.
Additionally, for Phase-$1$ of supervised learning, we employ Mixup~\cite{zhang2018mixup} and CutMix~\cite{9008296} with the respective mixing coefficients of $0.8$ and $1.0$, respectively following the work~\cite{fan-iclr2022}.
The drop path is set with a rate of $0.1$ and label smoothing at a rate of $0.1$.
Training is conducted using NVIDIA V100 GPUs. For Phase $1$ and Phase $2$ (refer Sec.~\ref{subsec:framework}), we conduct experiments for 25 and 50 epochs, respectively, across all datasets.
We used AdamW optimizer with a weight decay of $0.05$ along with a cosine learning rate scheduler having a base learning rate of $0.0001$.
For training, we utilize a mini-batch having $L_b$ labeled samples  and $\mu \times L_b$ unlabeled samples.
We set the values of  $\mu$ and $\tau$ (refer Eqn.~\ref{eq:instance}) to $4$ and $0.5$ respectively while the $\gamma$ and $\beta$ values (refer Eqn.~\ref{eq:total_loss}) are set to $0.6$ and $1$, respectively, unless otherwise stated.
During inference, only the primary pathway is used with 8 frame super images as inputs.
Additional dataset details and experimental results are provided in the Appendix.

\begin{table}[t]
    \centering  
    \renewcommand{\arraystretch}{1.5}
    \caption{\textbf{Comparisons of performance on UCF-101 and Kinetics-400.} We report the top-$1$ accuracy of \ours \ along with different state-of-the-art approaches over two different labeled data proportions. As observed, \ours \ achieves the best performance in both the datasets over the different amounts of labeled data. Also, \ours \: significantly reduces the compute needs as it uses much less model parameters. The values in \textbf{bold} denote the highest top-$1$ accuracy across all the specified models, respectively.}  
    \label{tab:ucfandk400}  
    \resizebox{0.95\textwidth}{!}{
    \begin{tabular}{lcccccccc}
    
    \hline 
    \multirow{2}{*}{ Method } & \multirow{2}{*}{ Backbone } & \multirow{2}{*}{ Params } & \multirow{2}{*}{ w \textbackslash ImgNet }  & \multicolumn{2}{c}{ UCF-101 } & \multicolumn{2}{c}{ Kinetics-400 }  \\
    \cline { 5 - 8 } & &  & & $1 \%$ & $10 \%$ & $1 \%$ & $10 \%$ \\
    \hline 
    \multirow{1}{*}{Supervised} & 3D-ResNet-50 &&  $\checkmark$  & $6.5$ & $32.4$ & $4.4$ & $36.2$ \\
    \hline
    FixMatch (NeurIPS'20) \cite{sohn2020fixmatch} & SlowFast-R50 & & $\checkmark$  & $16.1$ & $55.1$ & $10.1$ & $49.4$ \\
    VideoSSL(WACV'21) \cite{Jing_2021_WACV}  & 3D-ResNet-18 && $\checkmark$  & - & $42.0$ & - & $33.8$ \\

    TCL (CVPR'21) \cite{singh2021semisupervised}  & TSM-ResNet-18 & & & - & - & $8.5$ & - \\
    ActorCutMix (CVIU'21) \cite{zou2022learning} & R(2+1)D-34 && $\checkmark$ & - & $53.0$ & $9.02$ & $33.8$ \\
    
    MvPL (ICCV'21) \cite{xiong2021multiview} & 3D-ResNet-50 & && $22.8$ & $80.5$ & $17.0$ & $58.2$ \\
    CMPL (CVPR'22) \cite{xu2022crossmodel} & R50+R50-1/4 && $\checkmark$  & $25.1$ & $79.1$ & $17.6$ & $58.4$ \\
    LTG (CVPR'22) \cite{xiao2022learning} & 3D ResNet 18 & &  & - & $62.4$ & $9.8$ & $43.8$\\
    TACL(TCSVT'22) \cite{9904603}  & 3D-ResNet-50 && $\checkmark$  & - & $55.6$ & - & - \\
    L2A (ECCV'22) \cite{gowda2022learn2augment} & 3D-ResNet-18 && $\checkmark$ & - & $60.1$ & - & - \\
    \hline
    SVFormer-S(CVPR'23) \cite{svformer} & TimeSformer-S* & $81$M &$\checkmark$ & $31.4$ & $79.1$ & $32.6$ & $61.6$ \\
    \ours-S(Ours) & Swin-S & $49$M & $\checkmark$  & $\textbf{37.9}$ & $\textbf{81.8}$ & $\textbf{36.7}$ & $\textbf{64.1}$\\
    \hdashline
    SVFormer-B(CVPR'23) \cite{svformer} & TimeSformer(Default) & $121$M & $\checkmark$  & ${46.3}$ & {$86.7$} & $\textbf{49.1}$ & $\textbf{69.4}$ \\
    \ours-B(Ours) & Swin-B & $87$M & $\checkmark$  & $\textbf{47.0}$ & $\textbf{87.1}$ & {$39.0$} & {$66.5$} \\
    \hline
    \end{tabular}
    }
    \renewcommand{\arraystretch}{1}
    \renewcommand{\footnotesize}{\fontsize{5pt}{5pt}\selectfont\raggedright} 
    \footnotesize{*Following SVFormer, as TimeSformer only have ViT-B models, SVFormer-S model is implemented from ViT-S.}
\vspace{-0.7cm}
\end{table}

\begin{table}
    \centering  
    \renewcommand{\arraystretch}{1.5}
    \caption{\small\textbf{Comparisons with state-of-the-art methods on HMDB51.} We report the top-1 accuracy of \ours \: along with different state-of-the-art baselines over three different labeled data proportions. As observed, \ours \: achieves the best performance over the different amounts of labeled data.}  
    \label{tab:hmdb51}  
    \resizebox{0.7\textwidth}{!}{
    \begin{tabular}{lc@{\hspace{1cm}}c@{\hspace{0.5cm}}c@{\hspace{0.5cm}}c}
    
    \hline 
    Method  &  Backbone  & $40 \%$ & $50 \%$ & $60 \%$ \\
    
    \hline 
    
    VideoSSL \cite{Jing_2021_WACV}  & 3D-R18 & $32.7$ & $36.2$ & $37.0$ \\
    ActorCutMix  \cite{zou2022learning} & R(2+1)D-34  & $32.9$ & $38.2$ & $38.9$ \\
    MvPL  \cite{xiong2021multiview} & 3D-R18 & $30.5$ & $33.9$  & $35.8$\\
    LTG  \cite{xiao2022learning} & 3D-R18 & $46.5$ & $48.4$ & $49.7$\\
    TACL \cite{9904603}  & 3D-R18 & $38.7$ & $40.2$ & $41.7$\\
    L2A \cite{gowda2022learn2augment} & 3D-R18 & $42.1$ & $46.3$ & $47.1$ \\
    \hline
    SVFormer-S  \cite{svformer} &  TimeSformer-S* & $56.2$ & $58.2$ & $59.7$ \\
    \ours-S(Ours) & Swin-S & $\textbf{60.6}$ & $\textbf{62.6}$ & $\textbf{65.1}$ \\
    \hdashline
    SVFormer-B \cite{svformer} & TimeSformer(Default) & $61.6$  & $64.4$ & $68.2$ \\
    \ours-B(Ours) & Swin-B  & $\textbf{63.4}$ & $\textbf{65.5}$\ & $\textbf{68.2}$\\
    \hline
    \end{tabular}
    }

    \renewcommand{\arraystretch}{1}
        \renewcommand{\footnotesize}{\fontsize{5pt}{5pt}\selectfont\raggedright}
    \footnotesize{*Following standard procedure of SVFormer, as TimeSformer only have ViT-B models, SVFormer-S model is implemented from ViT-S.} 
    \vspace{-2\baselineskip}

\end{table}
\vspace{-3mm}
\subsection{Main Results}
\label{sub:main_results}
We evaluate our \ours's performance on three benchmark datasets: Kinetics-$400$~\cite{kay2017ics}, UCF-$101$~\cite{soomro2012ucf101}, and HMDB-$51$~\cite{hmdb}.
As can be seen in Tables~\ref{tab:ucfandk400} and \ref{tab:hmdb51} our \ours-S model consistently achieves superior performance on these datasets compared to the state-of-the-art, SVFormer-S~\cite{svformer}, while utilizing significantly less model parameters.
When using only $1\%$ labeled data, our \ours-S outperforms SVFormer-S by a significant margin of $6.5\%$ on UCF-$101$ and $4.1\%$ on Kinetics-$400$ (Table~\ref{tab:ucfandk400}).
This trend continues at the $10\%$ labeled data setting, with improvements of $2.7\%$ and $2.5\%$ for UCF-$101$ and Kinetics-$400$, respectively. Similarly, on HMDB-$51$ (Table\ref{tab:hmdb51}), \ours-S surpasses SVFormer-S by a substantial margin across all scenarios using different portions of labeled data ($40\%$, $50\%$, and $60\%$).

Our larger model, \ours-B, is at par with SVFormer-B while maintaining a significant parameter efficiency.
In UCF-$101$, \ours-B performs slightly better in both $1\%$ and $10\%$ labeled data settings compared to SVFormer-B.
Although our model is only second to SVFormer-B Kinetics-$400$, it has a significantly lower number of parameters compared to SVFormer-B.
Specifically, our model utilizes $28\%$ fewer parameters than the highest-performing model.
On HMDB-$51$, \ours-B exhibits improvements of $1.8\%$ and $1.1\%$ in the $40\%$ and $50\%$ settings, respectively, while being at par at $60\%$.
These results highlight the effectiveness of \ours \ in achieving high accuracy 
with reduced model complexity.

\vspace{-3mm}
\subsection{Ablation Results}
\label{sub:ablations}
We perform several ablation studies on HMDB-$51$ dataset to evaluate the effectiveness of the different components of our \ours \ framework.
\vspace{-4mm}
\subsubsection{Impact of Group Contrastive Loss.}
\begin{wraptable}{!tr}{0.65\textwidth}
    \centering  
    \renewcommand{\arraystretch}{1.5} 
    \caption{\small\textbf{Ablation Studies on HMDB-51.} Numbers show top-1 accuracy with $40\%$ labeled data.
}  
    \label{tab:impact_gcl}  
   \vspace{-2mm}
    \resizebox{0.65\textwidth}{!}{
    \begin{tabular}{c@{\hspace{1cm}}c}
    
    \hline 
    Approch  &  Top-1 Accuracy \\
    \hline
    \ours-S \: w/o Group-Contrastive Loss & $60.1$ \\
    \ours-S \: w/ Pseudo-Label Consistency Loss & $58.5$ \\
    \hdashline
    \ours-S (Ours) &  $\textbf{60.6}$\\
    \hline
    \end{tabular}
    }
    \renewcommand{\arraystretch}{1}
\vspace{-2\baselineskip}

\end{wraptable}
The importance of group contrastive loss in capturing high-level action semantics is evident from an ablation where we exclude Group Contrastive Loss from our framework (refer Section~\ref{subsubsec:gcl}) leading to a drop in top-$1$ accuracy from $60.6\%$ to $60.1\%$ shown in Table~\ref{tab:impact_gcl}.
\vspace{-5mm}
\subsubsection{Importance of Contrastive Loss.}
To evaluate the efficacy of contrastive loss, we compared it to the pseudo-label consistency loss used in FixMatch~\cite{sohn2020fixmatch}. As shown in Table~\ref{tab:impact_gcl}, training with our contrastive loss achieved a significant improvement in top-1 accuracy on the HMDB-$51$ dataset, outperforming pseudo-label consistency loss by $2.13\%$. This finding highlights the effectiveness of contrastive loss in capturing meaningful relationships between unlabeled videos.
\vspace{-5mm}
\begin{table}
	\vspace{-4mm}
    \centering  
    \renewcommand{\arraystretch}{1.5}
    \caption{\small\textbf{Effect of different Frame rates in Super Image.} We evaluate on HMDB-51 and report the top-1 accuracy by changing the layout to 4 × 4 and 3 × 3, incorporating 16 and 8 frames in the respective pathways. This was achieved in two ways: by keeping the size of super image constant and by increasing the super image size. The effect of computation on increasing the super image size is depicted in column `Flops'.}  
    \label{tab:impact_frame_rate}  
    \resizebox{1\textwidth}{!}{
    \begin{tabular}{cc@{\hspace{0.3cm}}c@{\hspace{0.3cm}}c@{\hspace{0.3cm}}c@{\hspace{0.3cm}}c}
    \hline 
    Model  &  \#Frames in Fast SI* & \#Frames in Slow SI*& SIze of SI* & Flops(G) & Top-1 Accuracy \\
    \hline
    \ours-S & 4 & 3 & 576 & 61 & $59.4$ \\

    \ours-S & 4 & 3 & 768 & 109 & $61.9$ \\
    \hdashline
    \ours-S(Ours) & 3 & 2 & 576 & 61 & $60.6$\\
    \hline
    \end{tabular}
    }

    \renewcommand{\arraystretch}{1}
    \renewcommand{\footnotesize}{\fontsize{5pt}{5pt}\selectfont\raggedright}
    \footnotesize{*SI=Super Image}
\vspace{-0.7cm}
\end{table}
\vspace{-3mm}
\subsubsection{Ablations on Super Images.}

\vspace{-8mm}
\begin{wraptable}{!tr}{0.6\textwidth}
\vspace{-7mm}
    \centering  
    \renewcommand{\arraystretch}{1.5}
    \caption{\textbf{Effect of temporal Order in Super Image.} We report the top-1 accuracy for HMDB-$51$ using $40\%$ labeled data with reverse, random and normal order.}  
    \vspace{0.5\baselineskip}
    \label{tab:impact_super_frame_order}  
    \resizebox{0.6\textwidth}{!}{
    \begin{tabular}{c@{\hspace{1cm}}c}
    \hline 
    Approch  &  Top-1 Accuracy \\
    \hline
    \ours-S \: w/ reverse frame order  & $59.6$ \\
    \ours-S \: w/ random frame order  & $58.9$ \\
    \hdashline
    \ours-S (Ours) &  $\textbf{60.6}$\\
    \hline
    \end{tabular}
    }

    \renewcommand{\arraystretch}{1}
\vspace{-1.5\baselineskip}
\end{wraptable}
In Phase-2 of training, our framework employs super images of layouts $3 \times 3$ and $2 \times 2$ in the Primary and Secondary pathways, respectively. Here, we explore the impact of changing the layout to $4 \times 4$ and $3 \times 3$, incorporating 16 and 8 frames in the respective pathways (with one extra padded frame in each case).
This can be achieved in two ways: Firstly, by increasing the number of frames to 16 and 8 while keeping the super image size constant (\textit{i.e.}, $576\times 576$), thus reducing the size of each frame. 
Secondly, by maintaining the size of each frame constant and increasing the super image size (\textit{i.e.}, $768\times 768$).
The results in Table~\ref{tab:impact_frame_rate} demonstrate that maintaining the super image size constant while increasing the number of frames leads to information loss due to the reduction in frame size and reduces Top-1 accuracy (shown in the first row of Table~\ref{tab:impact_frame_rate}).
Conversely, in the second experiment, increasing the super image size enhances the Top-1 accuracy, albeit at the expense of increased computation (model flop going from 61G to 109G) (shown in the second row of Table~\ref{tab:impact_frame_rate}).

In another ablation experiment we examine the effect of the temporal order of frames in the super image.
We evaluate our model's performance across three input types with different temporal orders: normal, random, and reverse (refer Fig.~\ref{fig:super_image_order}).
In the normal order, the temporal sequence of the video remains unchanged during super image generation. Conversely, for the reverse and random orders, sampled frames are arranged in the super image in reverse and random orders, respectively.
Table~\ref{tab:impact_super_frame_order} showcases the impact of altering the input order on the performance of the HMDB-51 dataset.
These findings exemplifies the significance of having normal input frame order in model learning for action recognition tasks.

\begin{figure*}[!h]
  \centering
  \vspace{-6mm}
  \includegraphics[width=0.9\textwidth]{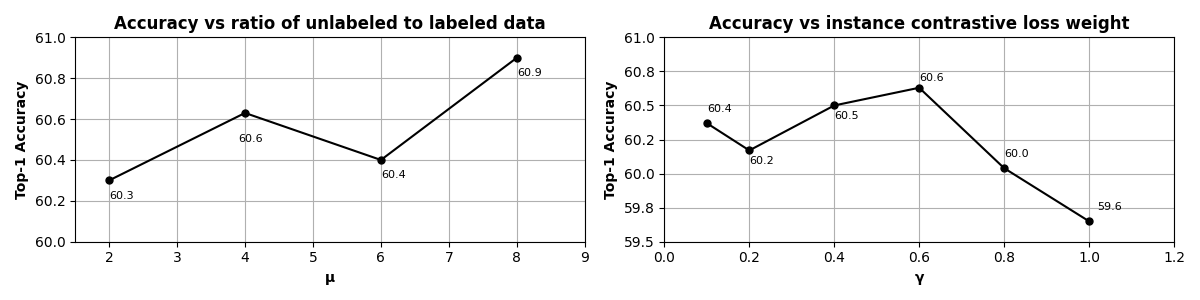}
  \vspace{-2mm}
  \caption{\small Effect of hyperparameters on HMDB51, (Left) Varying the ratio of unlabeled data to the labeled data ($\mu$),
(Right) Varying the instance-contrastive loss weight ($\gamma$)
  \label{fig:mu_and_gamma}
}
\vspace{-4mm}
\end{figure*}
\vspace{-7mm}
\subsubsection{Impact of Hyperparameters.}
\label{subsub:impact_hyperparams}
We investigate the impact of the ratio of unlabeled data to labeled data ($\mu$) and note that setting $\mu$ to ${2,4,6,8}$ with a fixed $\gamma$ $= 0.6$ yields similar outcomes on HMDB-51 (see Fig.~\ref{fig:mu_and_gamma}).
However, given the high computational demands associated with scaling $\mu$, we opt to set it to $4$ across all experiments to strike a balance between efficiency and accuracy in our model.
Additionally, we observe that the weight of the instance-contrastive loss ($\gamma$) influences performance in semi-supervised learning.
We check for $\gamma$ by setting it to ${0.1, 0.2, 0.4, 0.8, 1}$.
We find that $\gamma = 0.6$ gives the best result and we used this value of $\gamma$ throughout in our experiments.
\vspace{-3mm}

\begin{wraptable}{h}{0.7\textwidth}
    \centering  
    \renewcommand{\arraystretch}{1.5}
\vspace{-1mm}
    \caption{\small\textbf{Comparison of performance on UCF-101, HMDB-51 and Kinetics-400 using Swin-L backbone} We report the top-$1$ accuracy of \ours \ along with state-of-the-art approach SVFormer-B over different labeled data proportions. As observed, \ours \ achieves the best performance in all the datasets over different amounts of labeled data barring Kinetics-400 with 1\% labeled data.}
    \label{tab:main_sifarL}  
    \resizebox{0.7\textwidth}{!}{
    \begin{tabular}{lccc|ccc|cc}
    
    \hline 
    \multirow{2}{*}{ Method }  & \multirow{2}{*}{ Params }  & \multicolumn{2}{c}{ UCF-101 } & \multicolumn{3}{c}{ HMDB-51 } & \multicolumn{2}{c}{ K-400 }  \\
    \cline { 3 - 9 } & & $1 \%$ & $10 \%$ & $40 \%$ & $50 \%$ & $60 \%$ & $1 \%$ & $10 \%$\\
    \hline
    SVFormer-B(CVPR'23) \cite{svformer} & $121$M  & $ {46.3}$ & {$86.7$} & ${61.6}$ & ${64.4}$ & ${68.2}$ & ${49.1}$ & ${69.4}$ \\
    \ours-B(Ours) & $87$M  & ${47.0}$ & ${87.1}$ & ${63.4}$ & {${65.5}$} & {${68.2}$} & ${39.0}$ & ${66.5}$ \\
    \ours-L(Ours) & $195$M  & ${47.0}$ & ${87.2}$ & {${64.4}$} & {${67.7}$} & {${68.6}$} & ${44.0}$ & {${70.1}$} \\
    \hline
    \end{tabular}
    }
    \renewcommand{\arraystretch}{1}
    \renewcommand{\footnotesize}{\fontsize{5pt}{5pt}\selectfont\raggedright} %
    \vspace{-8mm}
    
\end{wraptable}
\subsubsection{Larger backbones.}
\label{sec:larger_architecture}
We conducted additional experiments with a larger swin transformer backbone Swin-L resulting in, \ours-L, having approximately $195$M parameters. We report that the results for \ours-L which outperforms SVFormer-B (refer Table~\ref{tab:main_sifarL}).  \ours-L demonstrates superior performance across different percentages of labeled data in different datasets (barring the scenario when 1\% of Kinetics-400 data is used with labels). Note that \ours-L is also able to outperform SVFormer-B in Kinetics-$400$ 10\% labeled data setting which \ours-B was falling short (refer Table~\ref{tab:ucfandk400}).

\begin{wraptable}{!tr}{0.7\textwidth}
    \centering  
    \renewcommand{\arraystretch}{1.5}  

\vspace{-8mm}
    \caption{\small\textbf{Comparison of parameters and FLOPs}. We show the comparison of parameters and total operations between our \ours\ and next best method  SVFormer.}  
    \label{tab_main:eff_comp}  
    \resizebox{0.65\textwidth}{!}{
    \begin{tabular}{c@{\hspace{0.3cm}}c@{\hspace{0.3cm}}c@{\hspace{0.3cm}}c}
    
    \hline 
    Method  &  Backbone & Params(M) & FLops(G)  \\
    
    \hline

    SVFormer-B \cite{svformer} & TimeSformer(Default) & $121$ & $196$  \\
    \ours-B(Ours) & Swin-B & $\textbf{87}$ & $\textbf{106}$ \\
    \hline
    \end{tabular}
    }

\vspace{-8mm}
    \renewcommand{\arraystretch}{1}  

\end{wraptable}
\vspace{-6mm}
\subsubsection{Flops and Parameters Comparison.}
\label{sec:eff-comp}
In Table \ref{tab_main:eff_comp}, we compare the total number of trainable parameters and the total number of operations (FLOPs) between \ours\ and the next best approach, SVFormer \cite{svformer}. It is evident that \ours-B utilizes $28\%$ fewer parameters compared to SVFormer-B, and the flops count is notably reduced by $45.9\%$. Despite having fewer parameters and a lower flop count, our model surpasses the Top-1 Accuracy for UCF-$101$ and HMDB-$51$ across all labeled data setting (refer Table~\ref{tab:ucfandk400} and~\ref{tab:hmdb51}).

\color{black}

\vspace{-5mm}
\section{Conclusions}
\vspace{-5mm}
This paper presents a novel approach for semi-supervised action recognition in videos using $2$D image transformer.  
By condensing input frames into a single super image, our method provides a straightforward yet potent means of re-purposing image classifiers for action recognition significantly reducing the compute cost. Employing a temporal contrastive learning framework maximizes the similarity between encoded representations of unlabeled videos transformed into super images at varying speeds, while minimizing similarity between different videos. Through the utilization of contrastive loss and exploration of high-level action semantics within video groups, our approach greatly mitigates the need of video annotations.
Our work outperforms several competitive approaches across three standard benchmark datasets.
These findings magnifies the potential of super images and temporal contrastive learning in advancing video understanding tasks, warranting further exploration and research in this area.
\vspace{-3mm}
\section*{Acknowledgements}
\vspace{-3mm}
This work was partially supported by the SERB Grant CRG/2023/005010.
We also acknowledge support from the MIT-IBM Watson AI Lab, which—through MIT Satori Supercomputer, contributed the computational resources necessary to conduct the experiments in this paper.
\vspace{-3mm}

\bibliographystyle{splncs04}
\bibliography{references}

\newpage
\section*{Appendix}
\appendix

\section{Dataset-Details}
\label{sec:dataset-details}
\subsubsection{UCF-$101$}
The UCF-$101$ dataset \cite{soomro2012ucf101}, comprises a collection of $13,320$ videos sourced from YouTube, encompassing $101$ distinct action categories. These categories are organized into five main types: $1)$ Human-Object Interaction, $2)$ Body-Motion Only, $3)$ Human-Human Interaction, $4)$ Playing Musical Instruments, and $5)$ Sports. All videos within the dataset are having frame rate of $25fps$ and a resolution of $320 \times 240$ pixels. The average duration of the clips are approximately $7.2$ seconds. Further, the dataset is structured into $25$ groups, each consisting of $4$ to $7$ videos of an action class. The clips in a group share similar characteristics like background, viewpoint etc.
The dataset is publicly available at \url{https://www.crcv.ucf.edu/data/UCF101.php}

\subsubsection{HMDB-$51$} The HMDB-$51$ dataset \cite{hmdb} stand for \textbf{H}uman \textbf{M}otion \textbf{D}ata\textbf{B}ase focuses on human actions, consisting of $51$ distinct action categories, have at least $101$ clip of each class making a total of $6766$ video clips, gathered from multiple source. The frame rate in all the videos is fixed at $30fps$ however, The height of frames in the video is scaled to $240$ pixels and width is adjusted to maintain the aspect ratio. The action categories are grouped in five types $1)$ General facial actions $2)$ Facial actions with object manipulation $3)$ General body movements $4)$ Body movements with object interaction $5)$ Body movements for human interaction. The dataset is publicly available at \url{https://serre-lab.clps.brown.edu/resource/hmdb-a-large-human-motion-database/}

\subsubsection{Kinetics-$400$}
The Kinetics-$400$ dataset \cite{kay2017ics} is the largest dataset among the three that we have considered, consisting of $306,245$ videos across $400$ distinct action categories with each class having at least $400$ clips. The average length of the videos is $10$ seconds. The action categories are grouped as $1)$ Person Actions:  drawing, drinking, laughing etc $2)$ Person-Person Action: hugging, kissing,
shaking hands $3)$ Person Object Action :opening present, mowing lawn, washing dishes. The dataset is publicly available at \url{https://deepmind.com/research/open-source/kinetics}

\section{Comparison of inference time.}
\label{sec:comp_inf_time}
We have compared the inference times for the \ours-B, \ours-L, and SVFormer-B models, measuring the time taken to process each video in seconds. The results are summarized in the Table~\ref{tab:inference_time} below. As shown in the table, both \ours-B and \ours-L models have significantly lower inference times compared to SVFormer-B model while the max batch size at the time of inference shows the efficiency of our model. Specifically, \ours-B takes 0.1 seconds and \ours-L takes 0.2 seconds inference time per video, while SVFormer-B takes 0.5 seconds per video. These results highlights the efficiency of our proposed models in terms of inference time compared to the SOTA, SVFormer.

\begin{table}
    \centering  
    \vspace{-1\baselineskip}
    \caption{\textbf{Comparison of Inference Time on UCF-$\textbf{101}$ dataset for \ours-B, \ours-L and SVFormer-B model.} We report the maximum batch size at inference time for the mentioned three models along with the inference time taken per video measured in seconds.}
    \vspace{1.5\baselineskip}
    \label{tab:inference_time}  
     \resizebox{0.8\textwidth}{!}{
    \begin{tabular}{c@{\hspace{0.5cm}}c@{\hspace{0.5cm}}c}
    \hline 
    Method  & Max Batchsize &  Inference Time(secs.) \\
    \hline
    SVFormer-B & $100$ & $0.5$ \\
    \ours-B  & $150$ & $0.1$ \\
    \ours-L  & $120$ & $0.2$ \\
    \hline
    \end{tabular}}
    \vspace{1.5\baselineskip}
\end{table}

\begin{wrapfigure}{!tr}{0.45\textwidth}
  \centering
  \includegraphics[width=0.9\textwidth]{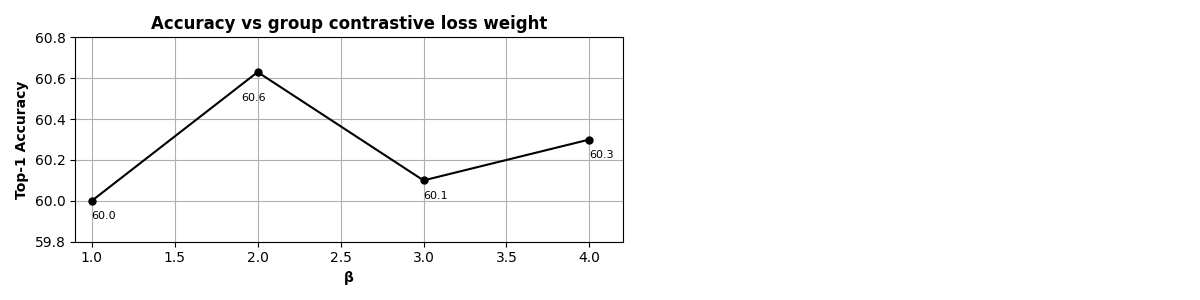}
  \vspace{-2\baselineskip}
  \caption{\small Effect of Hyperparameters on HMDB51, Varying the group-contrastive loss weight ($\beta$)}
  \label{fig:beta_abl}
\vspace{-1.7\baselineskip}
\end{wrapfigure}
\section{Effect of Group Contrastive Loss Weight, $\beta$}
\label{sec:add-exp}
Besides the ablation studies on different hyperparameters in fourth subsection of Section~$4$  of the main paper, here we also investigate the impact of varying the group contrastive loss weight, while keeping $\mu$ and $\gamma$ fixed at $4$ and $0.6$ respectively, on the HMDB-$51$ dataset with 40\% labeled data. Figure~\ref{fig:beta_abl} illustrates the performance changes corresponding to different values of $\beta$. It is apparent that varying $\beta$ influences the performance by up to $0.5\%$. Through hyperparameter tuning, we identified that $\beta = 2$ yields the optimal performance with the specified $\mu$ and $\gamma$ values.

\section{Feature visualization with t-SNE.}
\label{sec:tsne_plot}
\captionsetup[subfigure]{labelformat=empty}
\begin{figure}[!h]
\captionsetup[sub]{font=scriptsize}
    \centering
    
    \begin{subfigure}{0.3\columnwidth}
         \centering
         \includegraphics[width=\columnwidth]{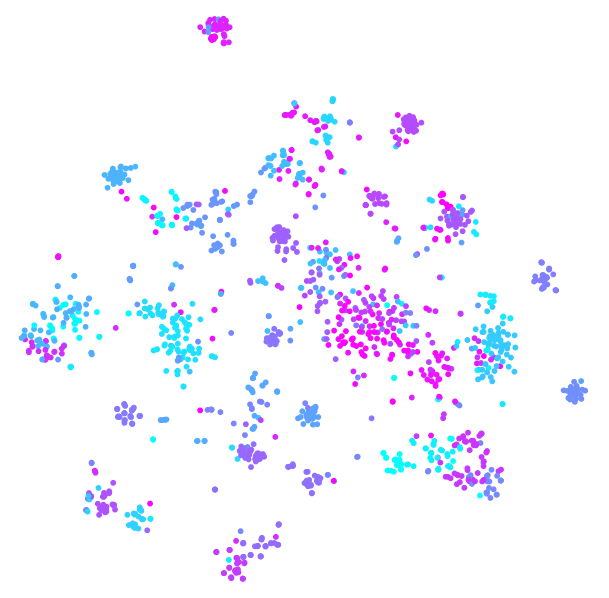}
         \caption{(a)}
    \end{subfigure}
    \begin{subfigure}{0.3\columnwidth}
         \centering
         \includegraphics[width=\columnwidth]{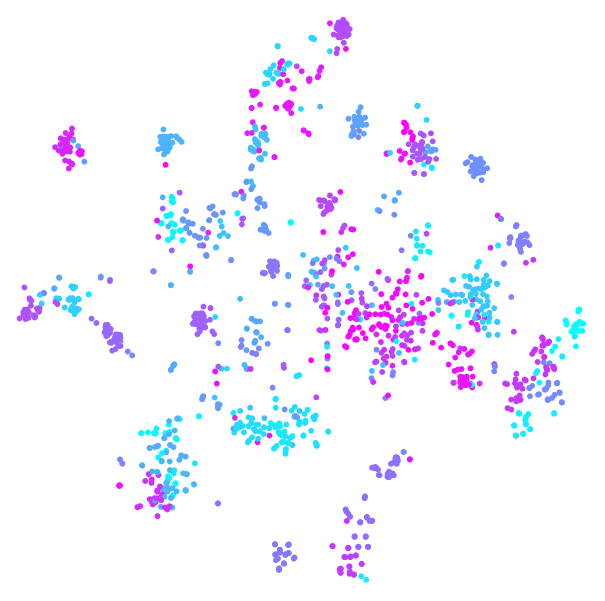}
         \caption{(b) }
    \end{subfigure}
    \begin{subfigure}{0.3\columnwidth}
         \centering
         \includegraphics[width=\columnwidth]{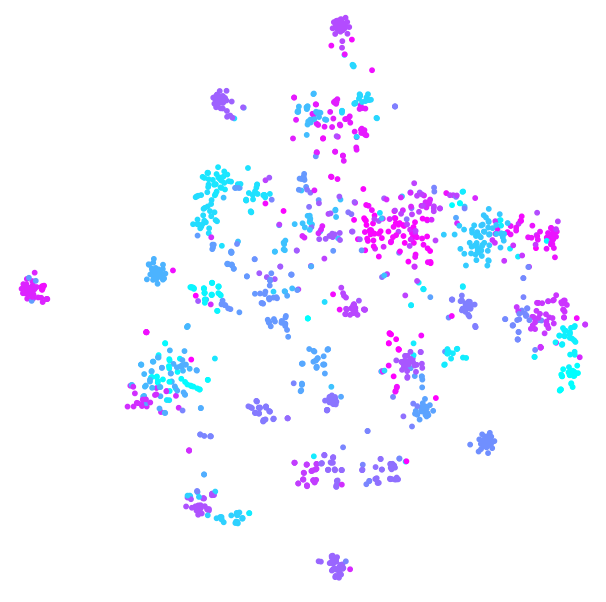}
         \caption{(c) }
    \end{subfigure}

    \begin{subfigure}{0.3\columnwidth}
         \centering
         \includegraphics[width=\columnwidth]{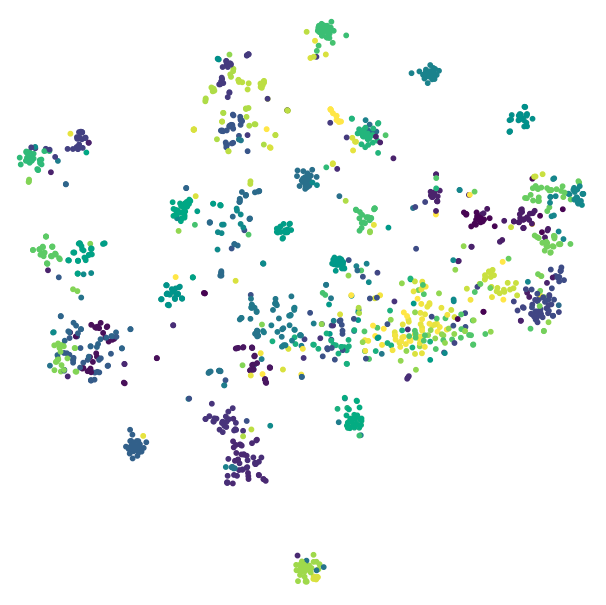}
         \caption{(d) }
    \end{subfigure}
    \begin{subfigure}{0.3\columnwidth}
         \centering
         \includegraphics[width=\columnwidth]{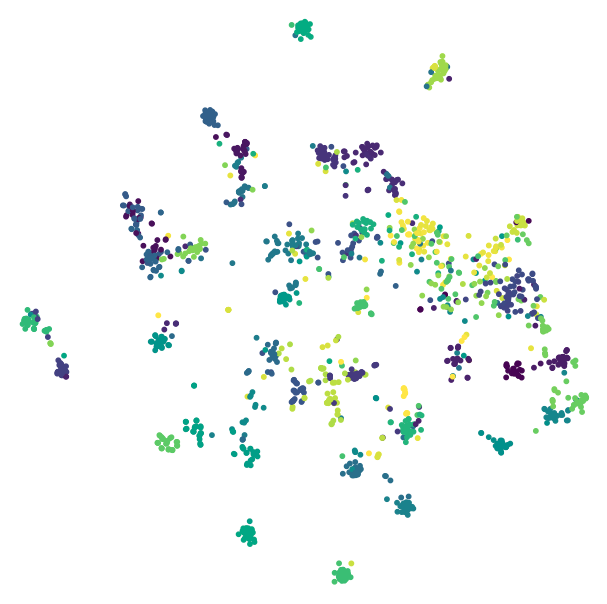}
         \caption{(e) }
    \end{subfigure}
    \begin{subfigure}{0.3\columnwidth}
         \centering
         \includegraphics[width=\columnwidth]{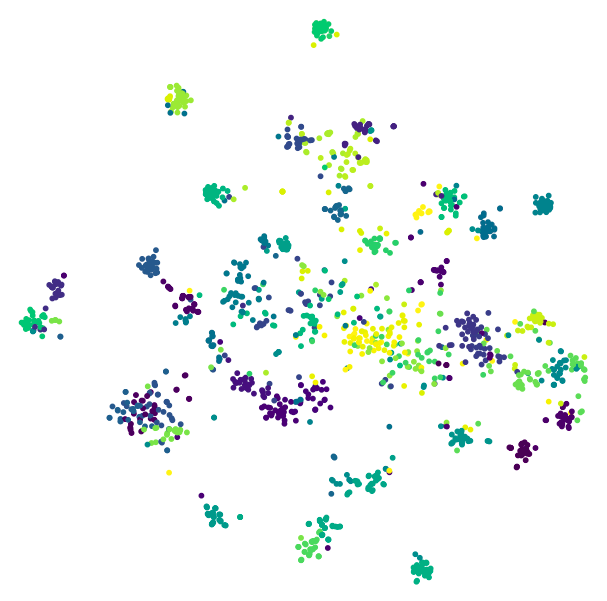}
         \caption{(f) }
    \end{subfigure}

        \caption{\textbf{Feature Visualization using t-SNE.} (a), (b), and (c) show the t-SNE plots for the HMDB-$51$ dataset at 40\%, 50\%, and 60\% labeled data setting for SVFormer-B model, while (d), (e), and (f) shows the same but for \ours-B model (Best viewed in color.)}
        \label{fig:supp_tsne_plots} \vspace{-4mm}
\end{figure}
\noindent The Fig.~\ref{fig:supp_tsne_plots} illustrates the clustering of target features for HMDB-$51$~\cite{hmdb} dataset under different labeled data settings ($40$\%, $50$\%, and $60$\%). The top three plots represent the SVFormer-B model, while the bottom three plots represent the \ours-B model. It is evident from the figure that the clustering for our \ours-B model is more consistent and discriminative compared to the SVFormer-B. This demonstrates the effectiveness of our model in learning discriminative features.

\section{Qualitative Examples}
\label{sec:qexp}
\begin{figure}[htb]
  \centering
  \setlength{\fboxrule}{2pt}
  \setlength{\fboxsep}{0pt}
  \captionsetup[subfigure]{labelformat=empty}
  \begin{subfigure}{0.23\linewidth}
    \centering\fbox{\includegraphics[width=\linewidth]{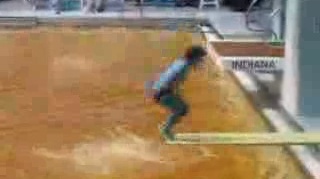}}
    \caption{\fontsize{1pt}{8pt}\selectfont \textbf{SVFormer:} \textcolor{red}{BreastStroke} \\ \textbf{SITAR:} \textcolor{myGreen}{Diving}}
  \end{subfigure}\hspace{2mm}
  \begin{subfigure}{0.23\linewidth}
    \centering\fbox{\includegraphics[width=\linewidth]{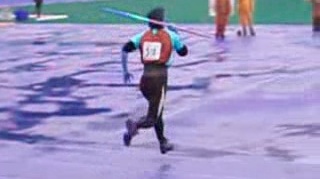}}
    \caption{\fontsize{1pt}{8pt}\selectfont \textbf{SVFormer:} \textcolor{red}{TennisSwing} \\ \textbf{SITAR:} \textcolor{myGreen}{JavelinThrow}}
  \end{subfigure}\hspace{2mm}
  \begin{subfigure}{0.23\linewidth}
    \centering\fbox{\includegraphics[width=\linewidth]{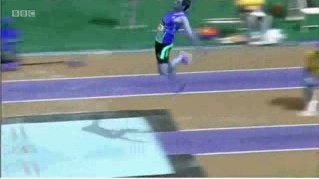}}
    \caption{\fontsize{1pt}{8pt}\selectfont \textbf{SVFormer:} \textcolor{red}{ParallelBars} \\ \textbf{SITAR:} \textcolor{myGreen}{LongJump}}
  \end{subfigure}\hspace{2mm}
  \begin{subfigure}{0.23\linewidth}
    \centering\fbox{\includegraphics[width=\linewidth]{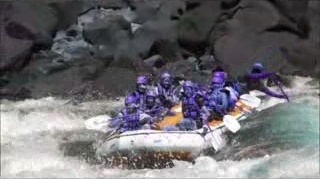}}
    \caption{\fontsize{1pt}{8pt}\selectfont \textbf{SVFormer:} \textcolor{red}{Kayaking} \\ \textbf{SITAR:} \textcolor{myGreen}{Rafting}}
  \end{subfigure}
  
  \vspace{0.5\baselineskip}
  \begin{subfigure}{0.23\linewidth}
    \centering\fbox{\includegraphics[width=\linewidth]{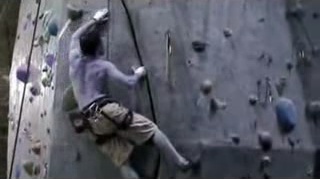}}
    \caption{\fontsize{1pt}{8pt}\selectfont \textbf{SVFormer:} \textcolor{red}{RopeClimbing} \\ \textbf{SITAR:} \textcolor{myGreen}{RockClimbing}}
  \end{subfigure}\hspace{2mm}
  \begin{subfigure}{0.23\linewidth}
    \centering\fbox{\includegraphics[width=\linewidth]{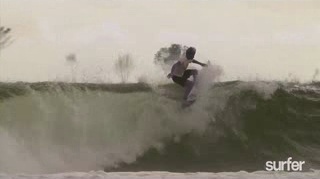}}
    \caption{\fontsize{1pt}{8pt}\selectfont \textbf{SVFormer:} \textcolor{red}{Skiing} \\ \textbf{SITAR:} \textcolor{myGreen}{Surfing}}
  \end{subfigure}\hspace{2mm}
  \begin{subfigure}{0.23\linewidth}
    \centering\fbox{\includegraphics[width=\linewidth]{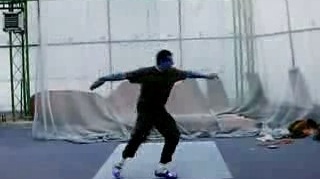}}
    \caption{\fontsize{1pt}{8pt}\selectfont \textbf{SVFormer:} \textcolor{red}{Yo Yo} \\ \textbf{SITAR:} \textcolor{myGreen}{ThrowDiscus}}
  \end{subfigure}\hspace{2mm}
  \begin{subfigure}{0.23\linewidth}
    \centering\fbox{\includegraphics[width=\linewidth]{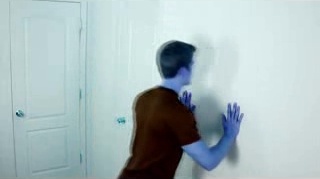}}
    \caption{\fontsize{1pt}{8pt}\selectfont \textbf{SVFormer:} \textcolor{red}{Pullups} \\ \textbf{SITAR:} \textcolor{myGreen}{WallPushups}}
  \end{subfigure}
  \caption{Qualitative examples comparing \ours\ with SVFormer \cite{svformer} on UCF-101 dataset trained using 1\% labeled
data. Both rows provide top-1 predictions using SVFormer and \ours\ respectively from top to bottom. As
observed,  the competing methods fail to classify the correct actions in most cases, our proposed approach, \ours\ is able to correctly recognize different action in this dataset. The predictions marked in green match the ground truth labels, whereas the red marked predictions are wrong. (Best viewed in color.)}
  \label{fig:qexp_ucf}
\end{figure}

\begin{figure}[h]
\setlength{\fboxrule}{2pt}
\setlength{\fboxsep}{0pt}
  \centering
  \captionsetup[subfigure]{labelformat=empty}
  \begin{subfigure}{0.23\textwidth}
    \centering\fbox{\includegraphics[width=\textwidth]{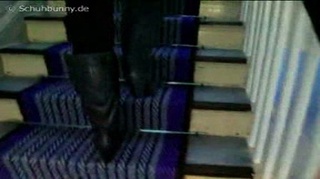}}
    \caption{\fontsize{1pt}{8pt}\selectfont  \textbf{SVFormer:} \textcolor{red}{walk} \\ \textbf{SITAR:} \textcolor{myGreen}{climb\_stairs}}
  \end{subfigure}\hspace{2mm}
  \begin{subfigure}{0.23\textwidth}
    \centering\fbox{\includegraphics[width=\textwidth]{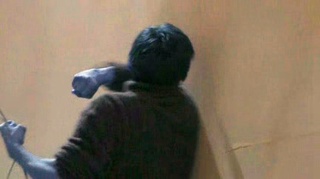}}
    \caption{\fontsize{1pt}{8pt}\selectfont \textbf{SVFormer:} \textcolor{red}{push} \\ \textbf{SITAR:} \textcolor{myGreen}{hit}}
  \end{subfigure}\hspace{2mm}
  \begin{subfigure}{0.23\textwidth}
    \centering\fbox{\includegraphics[width=\textwidth]{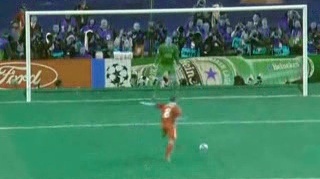}}
    \caption{\fontsize{1pt}{8pt}\selectfont \textbf{SVFormer:} \textcolor{red}{catch} \\ \textbf{SITAR:} \textcolor{myGreen}{kick\_ball}}
  \end{subfigure}\hspace{2mm}
  \begin{subfigure}{0.23\textwidth}
    \centering\fbox{\includegraphics[width=\textwidth]{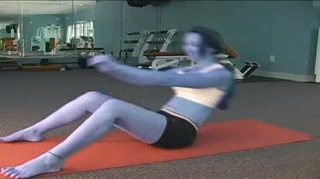}}
    \caption{\fontsize{1pt}{8pt}\selectfont \textbf{SVFormer:} \textcolor{red}{pick} \\ \textbf{SITAR:} \textcolor{myGreen}{situp}}
  \end{subfigure}
  
  \vspace{0.5\baselineskip}
  \begin{subfigure}{0.23\textwidth}
    \centering\fbox{\includegraphics[width=\textwidth]{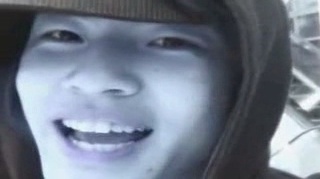}}
    \caption{\fontsize{1pt}{8pt}\selectfont \textbf{SVFormer:} \textcolor{red}{eat} \\ \textbf{SITAR:} \textcolor{myGreen}{smile}}
  \end{subfigure}\hspace{2mm}
  \begin{subfigure}{0.23\textwidth}
    \centering\fbox{\includegraphics[width=\textwidth]{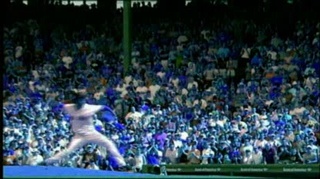}}
    \caption{\fontsize{1pt}{8pt}\selectfont \textbf{SVFormer:} \textcolor{red}{throw} \\ \textbf{SITAR:} \textcolor{myGreen}{swing\_baseball}}
  \end{subfigure}\hspace{2mm}
  \begin{subfigure}{0.23\textwidth}
    \centering\fbox{\includegraphics[width=\textwidth]{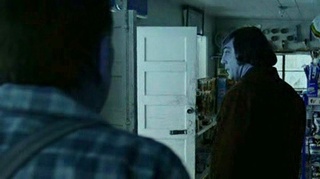}}
    \caption{\fontsize{1pt}{8pt}\selectfont \textbf{SVFormer:} \textcolor{red}{walk} \\ \textbf{SITAR:} \textcolor{myGreen}{turn}}
  \end{subfigure}\hspace{2mm}
  \begin{subfigure}{0.23\textwidth}
    \centering\fbox{\includegraphics[width=\textwidth]{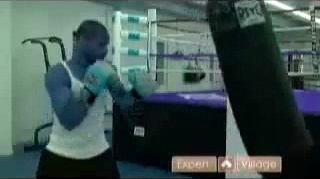}}
    \caption{\fontsize{1pt}{8pt}\selectfont \textbf{SVFormer:} \textcolor{red}{throw} \\ \textbf{SITAR:} \textcolor{myGreen}{punch}}
  \end{subfigure}
  \caption{Qualitative examples comparing \ours\ with SVFormer \cite{svformer} on HMDB51 dataset trained using 40\% labeled
data. Both rows provide top-1 predictions using SVFormer and \ours\ respectively from top to bottom. As observed,  the competing methods fail to classify the correct actions in most cases, our proposed approach, \ours\ is able to correctly recognize different action in this dataset. The predictions marked in green match the ground truth labels, whereas the red marked predictions are wrong. (Best viewed in color.)}
  \label{fig:qexp_hmdb}
\end{figure}

\begin{figure}[htb]
  \centering
\setlength{\fboxrule}{2pt}
\setlength{\fboxsep}{0pt}
  \captionsetup[subfigure]{labelformat=empty}
  \begin{subfigure}{0.23\linewidth}
    \centering\fbox{\includegraphics[width=\linewidth]{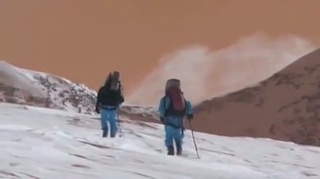}}
    \caption{\tiny \textbf{SVFormer:} \textcolor{red}{abseiling} \\ \textbf{SITAR:} \textcolor{myGreen}{ice climbing}}
  \end{subfigure}\hspace{2mm}
  \begin{subfigure}{0.23\linewidth}
    \centering\fbox{\includegraphics[width=\linewidth]{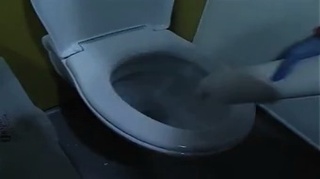}}
    \caption{\tiny \textbf{SVFormer:}\textcolor{red}{washing hands} \\ \textbf{SITAR:} \textcolor{myGreen}{cleaning toilet}}
  \end{subfigure}\hspace{2mm}
  \begin{subfigure}{0.23\linewidth}
    \centering\fbox{\includegraphics[width=\linewidth]{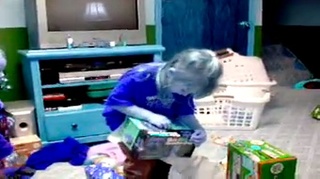}}
    \caption{\tiny \textbf{SVFormer:} \textcolor{red}{crawling baby} \\ \textbf{SITAR:} \textcolor{myGreen}{opening present}}
  \end{subfigure}\hspace{2mm}
  \begin{subfigure}{0.23\linewidth}
    \centering\fbox{\includegraphics[width=\linewidth]{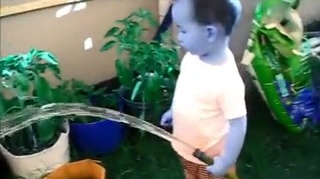}}
    \caption{\tiny \textbf{SVFormer:} \textcolor{red}{washing feet} \\ \textbf{SITAR:} \textcolor{myGreen}{watering plants}}
  \end{subfigure}
  
  \vspace{0.5\baselineskip}
  \begin{subfigure}{0.23\linewidth}
    \centering\fbox{\includegraphics[width=\linewidth]{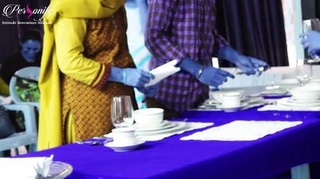}}
    \caption{\tiny \textbf{SVFormer:} \textcolor{red}{cooking egg} \\ \textbf{SITAR:} \textcolor{myGreen}{setting table}}
    \label{fig:sub5}
  \end{subfigure}\hspace{2mm}
  \begin{subfigure}{0.23\linewidth}
    \centering\fbox{\includegraphics[width=\linewidth]{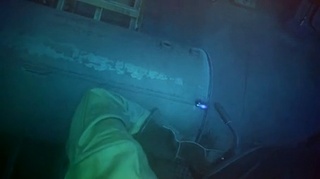}}
    \caption{\tiny \textbf{SVFormer:} \textcolor{red}{squat} \\ \textbf{SITAR:} \textcolor{myGreen}{welding}}
  \end{subfigure}\hspace{2mm}
  \begin{subfigure}{0.23\linewidth}
    \centering\fbox{\includegraphics[width=\linewidth]{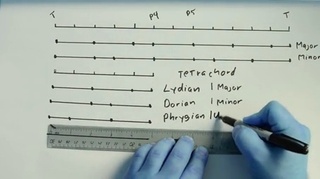}}
    \caption{\tiny \textbf{SVFormer:} \textcolor{red}{drawing} \\ \textbf{SITAR:} \textcolor{myGreen}{writing}}
  \end{subfigure}\hspace{2mm}
  \begin{subfigure}{0.23\linewidth}
    \centering\fbox{\includegraphics[width=\linewidth]{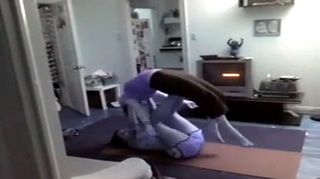}}
    \caption{\tiny \textbf{SVFormer:} \textcolor{red}{side kick} \\ \textbf{SITAR:} \textcolor{myGreen}{yoga}}
  \end{subfigure}
  \caption{Qualitative examples comparing \ours\ with SVFormer~\cite{svformer} on Kinetics-400 dataset trained using 1\% labeled data. Both rows provide top-1 predictions using SVFormer and \ours \ respectively from top to bottom. As observed,  the competing methods fail to classify the correct actions in most cases, our proposed approach, \ours \ \ is able to correctly recognize different action in this dataset. The predictions marked in green match the ground truth labels, whereas the red marked predictions are wrong. (Best viewed in color.)}
  \label{fig:qexp_k400}
\end{figure}

In Fig. \ref{fig:qexp_ucf}, \ref{fig:qexp_hmdb} and \ref{fig:qexp_k400} we provide diverse samples from $3$ different datasets of UCF-$101$, HMDB-$51$ and Kinetics-$400$ respectively. As observed, all of these samples are correctly classified by our proposed \ours\ while competing method failed, showing superiority of \ours\ over others.

\end{document}